\documentclass{bmvc2k}
\usepackage{multirow}
\usepackage{adjustbox}
\usepackage{xcolor,colortbl}
\usepackage{comment}
\usepackage{amsmath}
\usepackage{booktabs}
\usepackage{amssymb}

\title{Out-of-Distribution Detection from Small Training Sets using Bayesian Neural Network Classifiers}

\addauthor{Kevin Raina}{krain033@uottawa.ca}{1}
\addauthor{Tanya Schmah}{tschmah@uottawa.ca}{1}

\addinstitution{
 Department of Mathematics and Statistics\\
 University of Ottawa\\
 Ontario, CA
}

\runninghead{Raina,K., Schmah,T.}{OOD Detection using Bayesian Neural Nets}

\begin{document}

\maketitle

\begin{abstract}
Out-of-Distribution (OOD) detection is critical to AI reliability and safety, yet in many practical settings, only a limited amount of training data is available. Bayesian Neural Networks (BNNs) are a promising class of model on which to base OOD detection, because they explicitly represent epistemic (i.e. model) uncertainty.
In the small training data regime, BNNs are especially valuable because they can incorporate prior model information.
 We introduce a new family of Bayesian posthoc OOD scores based on expected logit vectors, and compare 5 Bayesian and 4 deterministic posthoc OOD scores.   Experiments on MNIST and CIFAR-10 In-Distributions, with 5000 training samples or less, show that the Bayesian methods outperform corresponding deterministic methods.

\end{abstract}

\section{Introduction}
\label{sec:intro}

The aim of ``Out-of-Distribution'' (OOD) detection is:
given a set of ``In-Distribution'' (ID) training samples, detect test samples that are OOD, i.e. that were not generated by the ID process. 
If the ID samples are considered ``normal'', then OOD samples are ``anomalies'', and their detection is an important problem in many applications in fields including medicine, science and cybersecurity. 
In this classic form, OOD detection is a one-class classification problem \cite{ruff2021unifying}.
One common approach to this problem is to build a generative model of the In-Distribution and then classify 
a test sample as OOD if its likelihood falls below a certain threshold \cite{bishop1994novelty}.
However there are many other approaches that do not rely on a generative model, such as: one-class SVM and variants \cite{ruff2018deep},
estimates of density, such as distance to $k^\textrm{th}$ nearest neighbour \cite{sun2022out}, and estimates of relative density such as the Local Outlier Factor \cite{Breunig2000lof}.

One application of OOD detection has attracted particular attention in the ML community: 
screening for samples that are OOD with respect to the training distribution of a discriminative model. 
The main aim is to improve reliability and AI safety by only applying models on ID data \cite{bishop1994novelty, nalisnick2019deep},
though see also \cite{Guerin_Delmas_Ferreira_Guiochet_2023}.
This has become a critical area of research, as machine learning models are increasingly deployed in real-world applications where encountering unfamiliar data is common. In this setting, we are given a set of \textit{labelled} ID samples, 
and our task is to detect OOD test samples.
This problem is what is usually meant by ``OOD Detection'' in the recent ML literature.
For nuanced discussions of variations of this and related problems, see
\cite{long2024rethinking, yang2024generalized}.

\begin{figure}
    \centering
    \includegraphics[scale=0.35]{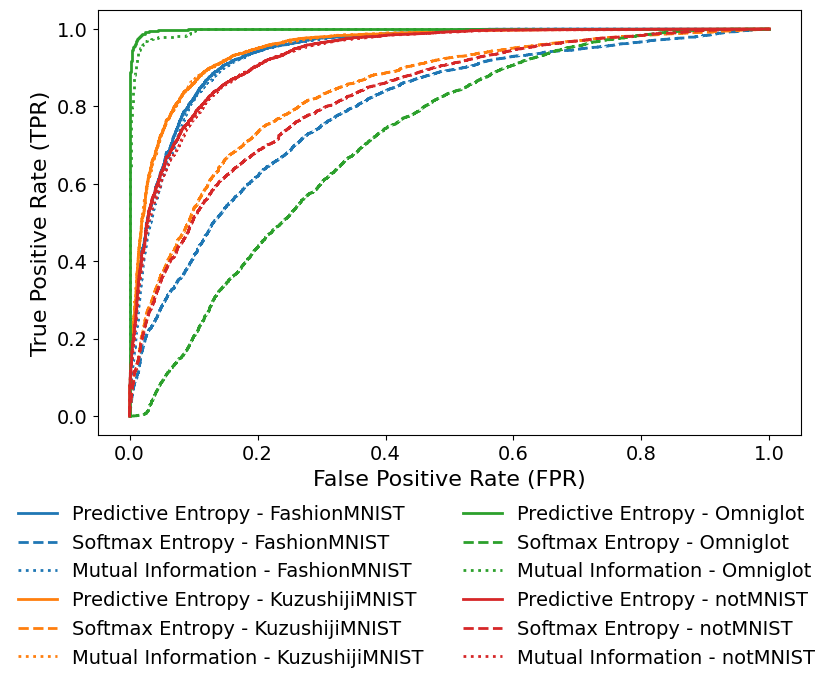}
    \caption{ROC curves comparing the OOD detection performance for MNIST experiments across four datasets and using three uncertainty estimation methods. Results are shown for a dataset size of 5000.}
    \label{fig:1}
\end{figure}

We consider the problem of OOD detection using Bayesian Neural Network (BNN) Classifiers, inspired by
insights in \cite{kendall2017uncertainties, smith2018understanding, kirsch2021PEharmful} regarding different sources of predictive uncertainty.
We return to this topic in detail in the Section \ref{sect:BayesianOOD}, however to summarise briefly: \textit{aleatoric uncertainty}
refers to noise in the true labels, 
while \textit{epistemic uncertainty} refers to uncertainty in the model specification.
One metric of the latter is the mutual information between labels and model parameters, and this has been
explored as an OOD detection score \cite{kendall2017uncertainties, rawat2017adversarial, smith2018understanding, minagawa2024out}.
In particular, the latter work showed that this method can exhibit high OOD detection performance
relative to an MNIST In-Distribution.
To illustrate the potential of Bayesian OOD methods, we show in Figure \ref{fig:1} the relative performance of two 
well-known Bayesian OOD scores (predictive entropy and mutual information)
and one deterministic one, softmax entropy, on an MNIST In-Distribution.
The Bayesian scores have consistently better performance. The performance of predictive entropy is particularly indicative
of a Bayesian advantage, since this method is a Bayesian version of softmax entropy \cite{malinin2018predictive}.


Our experiments focus on relatively small training datasets (up to 5000 examples), reflecting the real-world importance of OOD problems with limited data availability, in domains including medical imaging, industrial defect detection, fraud detection and remote sensing. 
Small-scale training datasets can limit classification performance, but this problem can addressed
by Bayesian learning, transfer learning and meta-learning. More relevant to the present work is that small training datasets can also hamper
our ability to distinguish between ID and OOD samples effectively, since they may lead to overfitted models that exhibit high confidence in their predictions, even for OOD samples.
We hypothesise that Bayesian methods are particularly well-suited to the small-training-set regime since we are able to leverage prior information in terms of parameter locations and distributions, while maintaining strong regularization capabilities.


Our main contributions are:  1) we compare Bayesian and deterministic versions of four post-hoc OOD scores, and find that the Bayesian versions have a consistent advantage, in experiments on MNIST and CIFAR-10 using small training sets,
2) we introduce a new OOD score based on k-NN in logit space that incorporates elements of both aleatoric and epistemic uncertainty, and
3) for BNNs, we find that OOD detection methods based on logit vectors tend to outperform the better-known methods of predictive entropy and mutual information.


The paper is organised as follows: we first review four deterministic post-hoc OOD scores; then we introduce BNNs and five Bayesian OOD scores, of which
four are Bayesian versions of the deterministic ones from the previous section (the fifth being mutual information).
Finally we report and discuss results of experiments based on two In-Distributions: MNIST and CIFAR-10, and four Out-Distributions for each.

\section{Deterministic Post-hoc OOD Scores}

In a standard neural network (NN) classifier, the output layer is a probability distribution on the labels, conditional on the input $\bf x$,
i.e. a vector of predicted probabilities $p(y_i | \bf x)$.
The penultimate layer contains one real-valued unit $z_i$, called a \emph{logit}, for every label, so it is a vector $\bf z$ of logits. The softmax function is
used to transform the logit vector into the probability vector, and for this reason the predicted probabilities are called \emph{softmax probabilities}.
Various OOD scores are based on logits or softmax probabilities. To apply the score for OOD detection, a threshold is determined 
on the basis of a set of ID validation samples, and then a test sample is classified as OOD if it is above the threshold (or below the threshold, for some tests). One of the simplest OOD scores is softmax entropy (SE), which is the entropy of the softmax probabilities, $H(\mathbf{x}) = -\sum_{i=1}^{K} p(y_i | \mathbf{x}) \log p(y_i | \mathbf{x}).$ A sample is classified as OOD if the SE is higher than a certain threshold.

\begin{figure*}[!h]
    \centering
    \includegraphics[scale=0.33]{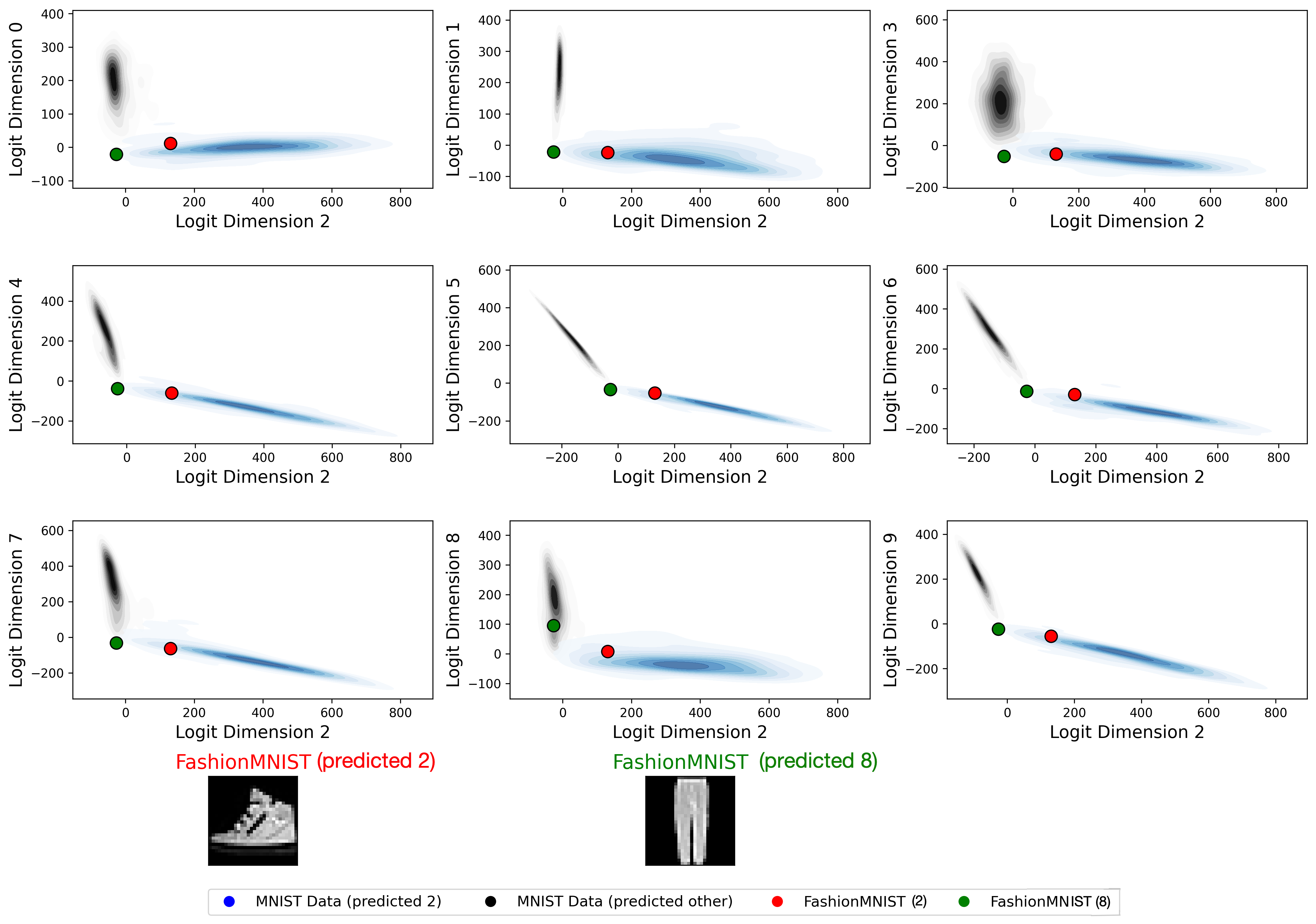}
    \caption{Density plots comparing the logits of MNIST test samples predicted 2 (blue), with remaining classes (black) and FashionMNIST samples (red and green) across different logit dimensions. The x-axis represents logit values for the selected dimension (2), while the y-axis varies across subplots. The two specific FashionMNIST samples used are illustrated below the main figure. Note in particular that the red point (predicted label 2) has a lower likelihood with respect to the multivariate distribution of the blue points (see in particular the logit 0 and 8 directions) than its likelihood with respect to the marginal distribution for logit 2.
    This illustrates that multivariate likelihood may better distinguish ID and OOD examples than the marginal likelihood of a single logit.
}
    \label{fig:2}
\end{figure*}

The Maximum Logit (Max Logit or ML) score is similar to the Maximum Softmax Probability (MSP), however for logits instead of softmax probabilities: $\textrm{MaxLogit}(\mathbf{x}) = \max_i z_i(\mathbf{x})$. A sample is classified as OOD if the Max Logit is lower than a certain threshold. The ML offers a unique advantage over the MSP, as the Softmax function can collapse informative structure in the logits, a recurring theme validated by our experiments.

\paragraph{k-NN Search in Logit Space.} Nearest neighbour search methods use the entire logit vectors $\bf z(x)$ instead of just the maximum logits.
It is similar in spirit to the Mahalanobis distance-based approaches in \cite{lee2018simple}. 
Figure \ref{fig:2} illustrates the possible value of considering entire logit vectors, with an example based on
an MNIST In-Distribution. As noted in the caption, multivariate likelihood may better distinguish ID and OOD examples than the marginal likelihood of a single logit $z_i(x)$.

In the logit space, we utilize the k-NN distance to compare the logits of a given input with those from the nearest ID training samples, $d_{\text{knn}}(\mathbf{x}) = \| \mathbf{z}(\mathbf{x}) - \mathbf{z}(\mathbf{x_k}) \|_2$ where \( \mathbf{z}(\mathbf{x_k}) \) represents the logit vector of the kth nearest neighbour, and \(\|\cdot\|_2\) denotes the Euclidean distance. This method follows a similar approach to Sun et al. \cite{sun2022out} and avoids the use of Mahalanobis distance, focusing on the Euclidean metric for simplicity. The standard k-NN distance is related to density, and is a surrogate measure of epistemic uncertainty \cite{postels2020hidden}.


\paragraph{Class-Conditioned k-NN Search.} To incorporate class-specific information and account for aleatoric uncertainty into the k-NN distance, we propose an additional term that modifies the standard k-NN distance by conditioning on the class labels of the neighbours. For each input \( \mathbf{x} \), we compute the k-NN distance within each class, restricted to neighbours that belong exclusively to that class. Specifically, for a given class \( c \), the class-conditioned k-NN distance is computed as the distance to the \( k \)-th nearest neighbour within class \( c \): $d_{\text{class}}(\mathbf{x}, c) = \| \mathbf{z}(\mathbf{x}) - \mathbf{z}(\mathbf{x_k}^{C_c}) \|_2$ where \( C_c \) is the set of training examples from class \( c \), and \( \mathbf{x_k}^{C_c} \) denotes the \( k \)-th nearest neighbour from within class \( C_c \). 
Next, we take the minimum distance across all classes and subtract the average k-NN distance to the remaining classes:

\begin{align*}
d_{\text{knn, conditioned}}(\mathbf{x}) &= d_{\text{knn}}(\mathbf{x}) + \min_c \left( \| \mathbf{z}(\mathbf{x}) - \mathbf{z}(\mathbf{x_k}^{C_c}) \|_2 \right) \\
&\quad - \frac{1}{C-1} \sum_{c' \neq c} \| \mathbf{z}(\mathbf{x}) - \mathbf{z}(\mathbf{x_k}^{C_{c'}}) \|_2
\end{align*}

Intuitively, this term captures the idea that OOD inputs are less likely to be close to the closest class but also less distant from other classes. Moreover, this score incorporates both elements of aleatoric and epistemic uncertainty as it penalizes ambiguous inputs (those that are "close" to many classes), and unfamiliar inputs (those that are generally in less dense regions). The logic behind this formulation is that OOD data points tend to be less confidently associated with any particular class and exhibit a higher degree of similarity to multiple classes simultaneously. We refer to this score as the \textit{class-conditional k-NN score} with abbreviation kNN+ in Tables \ref{Table1} and \ref{Table2}. 

\section{Bayesian OOD Detection}

\subsection{Bayesian Neural Networks}
Bayesian Neural Networks (BNNs) provide a principled approach to incorporating uncertainty in neural networks. Unlike traditional neural networks that rely on point estimates of parameters, BNNs model uncertainty by maintaining a distribution over weights, making them more robust to overfitting and data shifts \citep{blundell2015weight}. By tuning the prior distribution or employing empirical Bayes techniques, BNNs can incorporate prior information to improve generalization, even in small-scale, no-inventory settings \footnote{We distinguish a no-inventory setting as one where no prior knowledge of model weights is readily available.}. This characteristic makes BNNs particularly useful for OOD detection, where uncertainty estimation is crucial for distinguishing In-Distribution and Out-of-Distribution samples \citep{ritter2018scalable}. Additionally, the Bayesian framework provides a natural mechanism for regularization, improving generalization performance in few-shot learning settings.

For many prior distributions \(p(\omega)\), the posterior distribution over the weights, \( p(\omega | \mathcal{D}) \), given a dataset \( \mathcal{D} = \{ (\mathbf{x}_i, y_i) \}_{i=1}^N \), is typically intractable due to the complex, high-dimensional nature of the model. Therefore, variational inference techniques are commonly employed to approximate the true posterior by optimizing a simpler distribution \( q(\omega) \) that minimizes the Kullback-Leibler (KL) divergence between the true and approximate posteriors \cite{jordan1999introduction}. In our work, we specifically incorporate mean-field approximation, for its relative ease of use, and ability for prior distribution tuning. Variational inference methods, such as the mean-field approximation, assume that the posterior can be approximated by a factorized distribution \( q(\omega) = \prod_{j=1}^M q(\omega_j) \). By minimizing the KL divergence, we approximate the true posterior with a tractable distribution, allowing for efficient sampling and uncertainty quantification in BNNs. The posterior predictive distribution is then obtained by marginalizing over the weights: $p(y | \mathbf{x}, \mathcal{D}) = \int p(y | \mathbf{x}, \omega) q(\omega) d\omega.$ This framework provides a principled way to account for uncertainty in both the model parameters and predictions, making BNNs an appealing choice for tasks such as Out-of-Distribution (OOD) detection \cite{gal2016dropout,nguyen2022out,wang2021bayesian,maddox2019simple,lakshminarayanan2017simple}.



\subsection{Bayesian OOD Scores}\label{sect:BayesianOOD}

We consider five post-hoc OOD scores based on a BNN classifier. Each involves marginalizing over the 
posterior distribution of model parameters $\omega$ given the training set $D$. We employ two baseline OOD scores: \textit{predictive entropy} (PE) and \textit{mutual information} (MI),
\cite{kendall2017uncertainties, smith2018understanding}. Both scores are used to quantify uncertainty, with predictive entropy measuring uncertainty in a model's predictions and mutual information capturing the expected information gain in the model parameters over class labels.

\paragraph{Predictive Entropy.} Predictive entropy is a measure of predictive uncertainty in the model’s output, given a test input \( \mathbf{x} \), and is commonly used for OOD detection.
For deterministic classifiers it is more commonly called softmax entropy.
In the context of BNNs, we use entropy of the posterior predictive distribution: $H[p(y | \mathbf{x}, \mathcal{D})] 
=- \sum_{y} p(y | \mathbf{x}, \mathcal{D}) \log p(y | \mathbf{x}, \mathcal{D})$.
Here \( p(y | \mathbf{x}, \mathcal{D}) \) is the posterior predictive distribution over classes. Since the true posterior is often unknown, this is computed by averaging the softmax probability vector over multiple Monte Carlo samples of the weights \( \omega \) from the approximate posterior \( q(\omega) \). High entropy suggests high uncertainty in the prediction, which can indicate that the input is from an OOD sample, where the model’s confidence is low. Predictive entropy represents a combination of \emph{aleatoric certainty}, i.e. the noise in the true labels,
and \emph{epistemic uncertainty}, which is uncertainty in the model parameters \cite{smith2018understanding}.

\paragraph{Mutual information.} This score is the mutual information (MI) between the model parameters $\omega$ and the data, which
quantifies the amount of information gained about the parameters when observing a labelled data point.
It equals the difference between the entropy of the predictive distribution and the expected entropy of the model’s outputs conditioned on the input: $I(\omega, y \mid \mathcal{D},\mathbf{x}) = H[p(y | \mathbf{x}, \mathcal{D})] 
- \mathbb{E}_{p(\omega \mid \mathcal{D})}[H[p(y | \mathbf{x}, \omega)]],
$ where \( H[p(y | \mathbf{x}, \mathcal{D})] \) is the predictive entropy, and \( \mathbb{E}_{p(\omega\mid \mathcal{D})}[H[p(y | \mathbf{x}, \omega)]] \) represents the expected entropy across the posterior distribution over the weights. High mutual information indicates that knowing the label of the input $\bf x$ would tell us a lot about the model parameters, which implies that the model is uncertain about how to label $\bf x$. Thus MI is a measure of epistemic uncertainty. High MI suggests that the input is OOD.

\paragraph{Bayesian adaptation of OOD scores.} Given any OOD score based on logit vectors, we consider a corresponding \textit{Bayesian variant} based on the
expected logits obtained by averaging over the posterior distribution of the model parameters. Specifically, for each test sample $\bf x$, the expected logit vector (ELV) is $\hat{\mathbf{z}}(x) = \mathbb{E}_{p(\omega|D)}\left[\bf z(\bf x; \omega)\right]$
For each of the OOD scores Max Logit, kNN and kNN+ the Bayesian version is defined in exactly the same way but based on the ELV $\hat{\mathbf{z}}(\mathbf{x})$ for every $\bf x$
(training or test).

\section{Experiments}


\paragraph{BNN Training} 

For MNIST experiments, we do not use pretraining. We train a Bayesian LeNet-5 (see supplementary section for more details) and, as suggested by Blundell et al. \cite{blundell2015weight}, we tune a scale mixture Gaussian prior with two components over the model weights i.e, 
$p(\omega) = \prod_j\pi\mathcal{N}(\omega_j;0,\sigma^2_1) + (1-\pi)\mathcal{N}(\omega_j;0,\sigma_2^2).$  
For comparison, we also train a standard LeNet-5 using traditional stochastic gradient descent (SGD) on the cross entropy loss, which can be seen as performing approximate maximum likelihood estimation (MLE) of the model weights.

In our CIFAR-10 \cite{krizhevsky2009learning} experiments we train a Bayesian ResNet18 with informative priors. We apply the MOPED method \cite{krishnan2020specifying} with $\delta=0.01$ by employing weights pre-trained on the ImageNet1K \cite{zhang2021understanding} dataset. This is an empirical Bayes method, with a diagonal Gaussian prior.
We apply the transfer learning derivative fine-tuning (FinT) on a ResNet18 with pretrained weights from the ImageNet1K dataset as well, as in the BNN case, for a method that holds an equivalent deterministic representation. In fine-tuning, the model weights are loaded from a previous task, and we apply a gradient descent step, usually with smaller learning rates.

\paragraph{Small Training Datasets.} 
Our aim is to evaluate OOD detection performance in the regime of limited amounts of training data. Thus we 
consider smaller training data sets than are typically used in classification of MNIST and CIFAR-10 images. For the
MNIST experiments, we consider three training dataset sizes: $500$, $1000$ and $5000$ randomly selected, with balanced classes. In CIFAR-10 experiments, we used $5000$ training samples in a similar sampling framework, as the classification task is relatively difficult. We provide statistical tests of normality and unimodality to better understand the logit data distribution in the supplementary material.

\paragraph{Experimental setting.} In the first set of experiments, MNIST is the In-Distribution with FashionMNIST \cite{xiao2017fashion} (fashion-product images), Omniglot \cite{lake2015human} (handwritten characters from 50 alphabets), KuzushijiMNIST \cite{clanuwat2018deep} (kanji characters) and notMNIST \cite{bulatov2011notmnist} (various fonts of letters A through I) as OOD. For the second set of experiments CIFAR-10 is the In-Distribution with SVHN \cite{netzer2011reading}(cropped house number plates), CIFAR-100 (superset of CIFAR-10), Places365 \cite{zhou2017places} (scene recognition) and Textures \cite{cimpoi2014describing} (texture and pattern images) as OOD. We compute the following metrics: (1) The area under the ROC curve (AUC-ROC↑), and (2) false-positive rate at $95$ percent true-positive rate (FPR95$\downarrow$). Positive are treated as OOD in our experiments. 
In all Bayesian scores, $M = 500$ models from the posterior distribution were sampled. For additional experimental details see the supplementary section.

\section{Experimental Results}

Table \ref{Table1} and \ref{Table2} display OOD detection performance metrics for our MNIST and CIFAR-10 experiments. The best overall method, judged by the highest average AUC-ROC across the four OOD datasets for the given ID dataset, is EL kNN+, which is the Bayesian variant of 
our class-conditional adapted k-NN score. This method also has the lowest average FPR95. For the informed prior CIFAR-10 experiments, EL ML—the Bayesian variant of the Maximum Logit score—achieved the best performance, with the highest AUC-ROC and lowest FPR95.


\subsection{MNIST Experiments}


\begin{table*}[h!]
\centering
\begin{adjustbox}{max width=\textwidth}
\begin{tabular}{cccccccccc}
\toprule
\multicolumn{2}{c}{MNIST} & \multicolumn{2}{c}{FashionMNIST} & \multicolumn{2}{c}{Omniglot} & \multicolumn{2}{c}{KMNIST} & \multicolumn{2}{c}{notMNIST}\\
\midrule
\textbf{Dataset size} & \textbf{Score} & AUC-ROC$\uparrow$ & FPR95$\downarrow$ & AUC-ROC$\uparrow$ & FPR95$\downarrow$ & AUC-ROC$\uparrow$ & FPR95$\downarrow$ & AUC-ROC$\uparrow$ & FPR95$\downarrow$ \\

 

\midrule
   \multirow{1}{*}{}& SE & $52.42$ & $94.32$ & $14.52$ & $97.80$ & $66.19$ & $84.06$ & $54.19$ & $94.58$ \\
   \multirow{1}{*}{}& MI & $81.58$ & $50.34$ & $93.88$ & $16.08$ & $86.09$ & $43.70$ & $85.18$ & $\mathbf{40.34}$ \\  
 \multirow{1}{*}{}{} & PE & $83.10$ &$51.02$& $99.06$ & $5.28$ & $75.31$ & $86.22$ & $85.73$ & $40.40$  \\
\multirow{1}{*}{N=500}& MLE ML  & $52.95$ & $44.28$ & $5.60$  & $98.74$ & $62.79$ & $84.78$ & $44.32$ & $98.40$ \\
\multirow{1}{*}{BNN Acc:92.56\%}& EL ML  & $\mathbf{91.69}$ & $35.78$ & $\mathbf{99.94}$  & $2.80$ & $85.95$ & $44.28$ & $79.41$ & $68.84$\\
\multirow{1}{*}{NN Acc:}84.20\%& MLE kNN  & $77.13$ & $70.02$ & $99.13$ & $\mathbf{1.80}$ & $77.13$ & $65.44$ & $\mathbf{90.41}$ & $41.66$ \\
\multirow{1}{*}{}& EL kNN & $80.68$ & $52.84$ & $88.34$  & $14.92$ & $83.09$ & $50.44$ & $86.10$ & $39.82$\\
\multirow{1}{*}{}& MLE kNN+& $79.22$ & $71.78$ & $98.64$ & $4.76    
 $ & $78.85$ & $63.20$ & $82.86$ & $66.86$ \\
\rowcolor{lightgray} \multirow{1}{*}{} &  EL kNN+ & $90.06$ & $\mathbf{31.56}$ & $97.50$  & $4.38$ & $\mathbf{88.26}$ & $\mathbf{37.62}$ &  $85.71$  & $48.08$\\

\midrule

  \multirow{1}{*}{}& SE & $68.02$ & $86.58$ & $85.81$ & $41.98$ & $72.98$ & $77.04$ & $66.23$ & $89.14$ \\
 \multirow{1}{*}{}& MI & $83.96$ & $58.28$ & $95.27$ & $17.74$ & $92.17$ & $27.26$ & $90.16$ & $\mathbf{30.52}$ \\
 \multirow{1}{*}{}{} & PE & $84.64$ &$58.46$& $99.70$ & $1.80$ & $92.17$ & $27.58$ & $\mathbf{90.63}$ & $30.86$  \\
\multirow{1}{*}{N=1000}& MLE ML  & $61.00$ & $84.48$ & $28.44$  & $86.56$ & $65.24$ & $81.68$ & $47.04$ & $95.38$ \\
\multirow{1}{*}{BNN Acc:95.22\%}& EL ML  & $\mathbf{94.91}$ & $\mathbf{18.48}$ & $\mathbf{99.97}$  & $\mathbf{0.10}$ & $91.14$ & $32.10$ & $88.08$ & $42.62$\\
\multirow{1}{*}{NN Acc:88.26\%}& MLE kNN  & $73.83$ & $72.44$ & $99.96$ & $0.20$ & $77.09$ & $61.92$ & $90.61$ & $40.32$ \\
\multirow{1}{*}{}& EL kNN & $80.02$ & $77.32$ & $91.89$  & $10.22$ & $87.58$ & $36.54$ & $86.55$ & $41.02$\\
 \multirow{1}{*}{}& MLE kNN+& $80.33$ & $64.78$ & $98.81$ & $6.20    
 $ & $80.19$ & $60.12$ & $88.72$ & $49.38$ \\
\rowcolor{lightgray} \multirow{1}{*}{}&  EL kNN+ & $93.31$ & $24.32$ & $99.73$  & $0.58$ & $\mathbf{92.60}$ & $\mathbf{23.68}$ &  $89.68$  & $32.24$\\

\midrule
  \multirow{1}{*}{}& SE & $79.25$ & $71.02$ & $71.69$ & $68.04$ & $84.01$ & $59.58$ & $82.47$ & $61.34$ \\
  \multirow{1}{*}{}& MI & $93.75$ & $21.46$ & $99.53$ & $1.70$ & $95.10$ & $19.92$ & $92.90$ & $26.86$ \\
 \multirow{1}{*}{}{} & PE & $94.22$ &$21.40$& $99.87$ & $0.56$ & $95.12$ & $19.92$ & $93.22$ & $26.04$  \\
\multirow{1}{*}{N=5000}& MLE ML   & $83.41$ & $53.86$ & $59.17$  & $69.52$ & $82.86$ & $63.86$ & $79.04$ & $68.10$ \\
\multirow{1}{*}{BNN Acc:97.92\%}& EL ML  & $\mathbf{98.18}$ & $\mathbf{7.70}$ & $\mathbf{99.99}$  & $\mathbf{0.02}$ & $94.47$ & $22.06$ & $92.92$ & $27.14$\\
\multirow{1}{*}{NN Acc:97.01\%}& MLE kNN  & $89.16$ & $33.04$ & $98.57$ & $5.86$ & $89.25$ & $35.02$ & $91.87$ & $31.88$ \\
\multirow{1}{*}{}& EL kNN & $91.61$ & $26.40$ & $95.18$  & $6.52$ & $92.22$ & $26.64$ & $90.90$ & $35.42$\\
 \multirow{1}{*}{}& MLE kNN+& $95.30$ & $15.94$ & $97.32$ & $10.24$ & $92.91$ & $26.90$ & $90.61$ & $40.32$ \\
 \rowcolor{lightgray} \multirow{1}{*}{}&  EL kNN+ & $97.86$ & $7.94$ & $97.66$  & $0.44$ & $\mathbf{95.71}$ & $\mathbf{15.80}$ & $\mathbf{94.39}$ & $\mathbf{20.78}$ \\

\midrule

\end{tabular}
\end{adjustbox}
\caption{OOD performance data for MNIST experiments. The best method (EL kNN+), which has simultaneously the lowest average FPR95 and highest average AUC-ROC, is shaded in gray. The best individual method, per experiment, is bolded to show individual variability. The BNN and NN classification accuracies are also reported along the first column.}
\label{Table1}
\end{table*}
\vspace{0cm}
On average we notice BNNs outperform traditional NNs in softmax entropy scores, maximum logit scores, and our adapted logit-space k-NN score in the small data setting. Predictive entropy and mutual information are decisively better than the softmax entropy counterpart in OOD detection on small datasets. Figure \ref{fig:1} supports this claim by plotting the ROC curves for the softmax scores on FashionMNIST detection. For $N=500$ Omniglot and notMNIST experiments the traditional k-NN score for MLE-trained NNs displays somewhat better detection performance compared to the Bayesian counterparts, however, this seems to be significantly reciprocated on FashionMNIST and KuzushijiMNIST. As the dataset size increases our class-conditional Bayesian k-NN score appears to bridge the gap in detection performance on Omniglot and notMNIST while remaining better on FashionMNIST and KuzushijiMNIST.   
For both Bayesian and deterministic NNs, the measures based on Logit space (kNN, kNN+, ML) tend to outperform those based on softmax probabilities (PE, ML, SE). The notMNIST dataset was, mainly for the lower sample sizes, an exception to this pattern. 


\begin{table*}[h!]
\centering
\begin{adjustbox}{max width=\textwidth}
\begin{tabular}{cccccccccc}
\toprule
\multicolumn{2}{c}{CIFAR-10} & \multicolumn{2}{c}{SVHN} & \multicolumn{2}{c}{CIFAR-100} & \multicolumn{2}{c}{Places365} & \multicolumn{2}{c}{Textures}\\
\midrule
\textbf{Dataset size} & \textbf{Score} & AUC-ROC$\uparrow$ & FPR95$\downarrow$ & AUC-ROC$\uparrow$ & FPR95$\downarrow$ & AUC-ROC$\uparrow$ & FPR95$\downarrow$ & AUC-ROC$\uparrow$ & FPR95$\downarrow$ \\

\midrule
   \multirow{1}{*}{}& SE (FinT) & $90.80$ & $34.64$ & $82.72$ & $54.16$ & $92.49$ & $30.10$ & $95.85$ & $16.86$ \\
\multirow{1}{*}{}& MI & $78.95$ & $38.48$ & $78.41$ & $54.16$ & $60.83$ & $52.60$ & $56.43$ & $54.88$ \\
 \multirow{1}{*}{}{} & PE & $91.08$ &$32.18$& $83.37$ & $53.52$ & $93.44$ & $29.42$ & $96.94$ & $16.18$  \\ 
\multirow{1}{*}{N=5000}& FinT ML  & $91.37$ & $32.64$ & $82.81$  & $56.86$ & $93.30$ & $25.40$ & $95.26$ & $18.36$ \\

\rowcolor{lightgray} \multirow{1}{*}{BNN Acc:87.95\%}& EL ML  & $\mathbf{92.95}$ & $27.74$ & $\mathbf{84.75}$  & $51.28$ & $\mathbf{95.05}$ & $\mathbf{22.20}$ & $\mathbf{97.58}$ & $\mathbf{10.88}$\\ 

\multirow{1}{*}{NN Acc: 86.13\%}& FinT kNN  & $82.47$ & $40.86$ & $77.97$ & $63.28$ & $82.29$ & $38.64$ & $84.47$ & $26.56$ \\

\multirow{1}{*}{}& EL kNN & $83.84$ & $31.18$ & $78.75$  & $57.04$ & $79.91$ & $39.06$ & $82.50$ & $29.18$\\
 \multirow{1}{*}{}& FinT kNN+& $91.49$ & $29.22$ & $83.15$ & $54.92    
 $ & $91.84$ & $26.18$ & $94.66$ & $16.62$ \\
 
\multirow{1}{*}{}& EL kNN+ & $92.19$ & $\mathbf{24.88}$ & $84.17$  & $\mathbf{49.78}$ & $92.12$ & $26.62$ &  $95.03$  & $17.40$\\

\midrule

\end{tabular}
\end{adjustbox}
\caption{OOD performance data for CIFAR-10 experimental setup. The best method (EL ML), is deemed by simultaneously having the lowest average FPR95 and highest average AUC-ROC over experiments are shaded in gray. The best individual method, per experiment, is bolded to show individual variability. The BNN and NN classification accuracies are also reported along the first column.}
\label{Table2}
\end{table*}

 \begin{figure*}[!h]
    \centering
        \begin{subfigure}{}
             \includegraphics[scale=0.40]{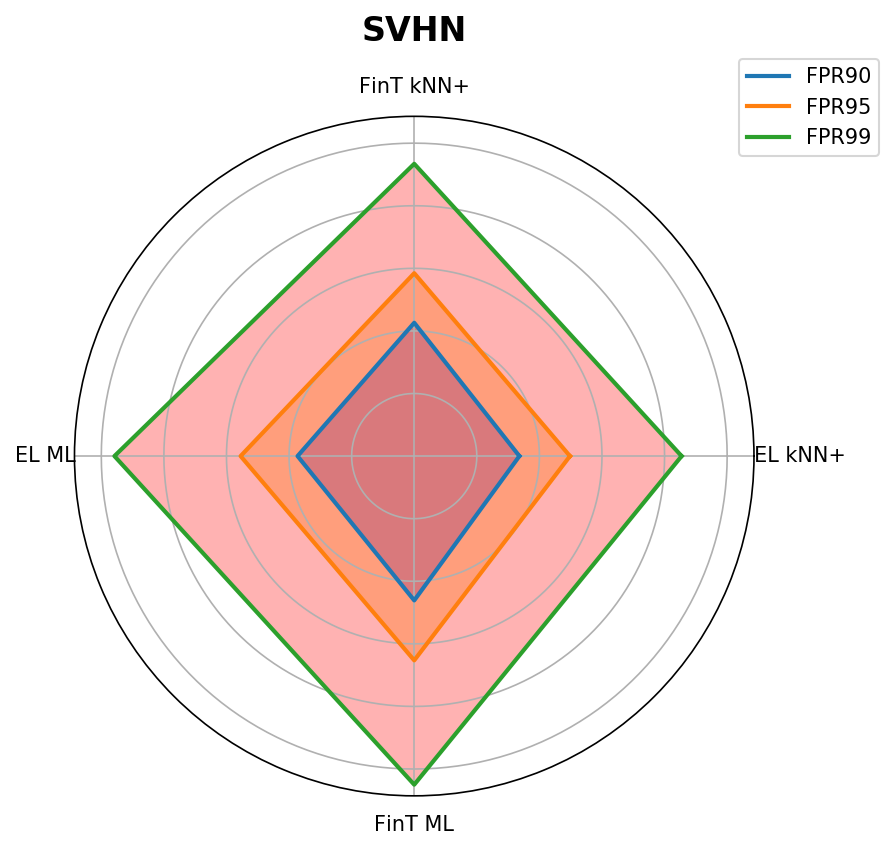}
        \end{subfigure}        
        \begin{subfigure}{}
          \includegraphics[scale=0.40]{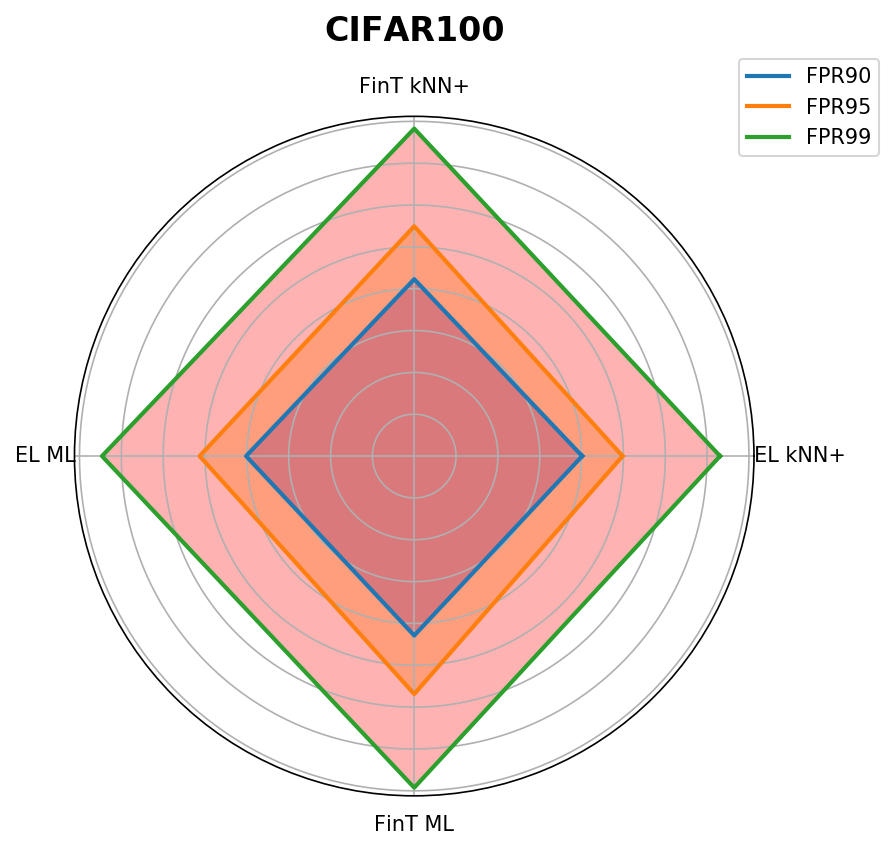}   
        \end{subfigure}
    
     \begin{subfigure}{}
      \includegraphics[scale=0.40]{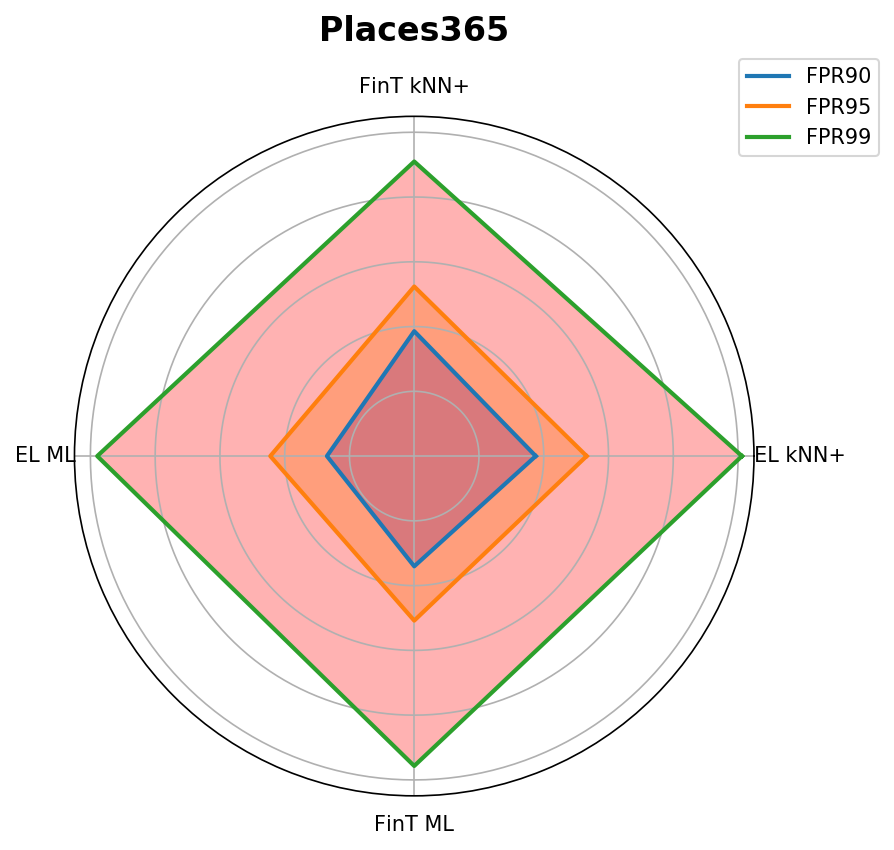}      
    \end{subfigure}
         \begin{subfigure}{}
      \includegraphics[scale=0.40]{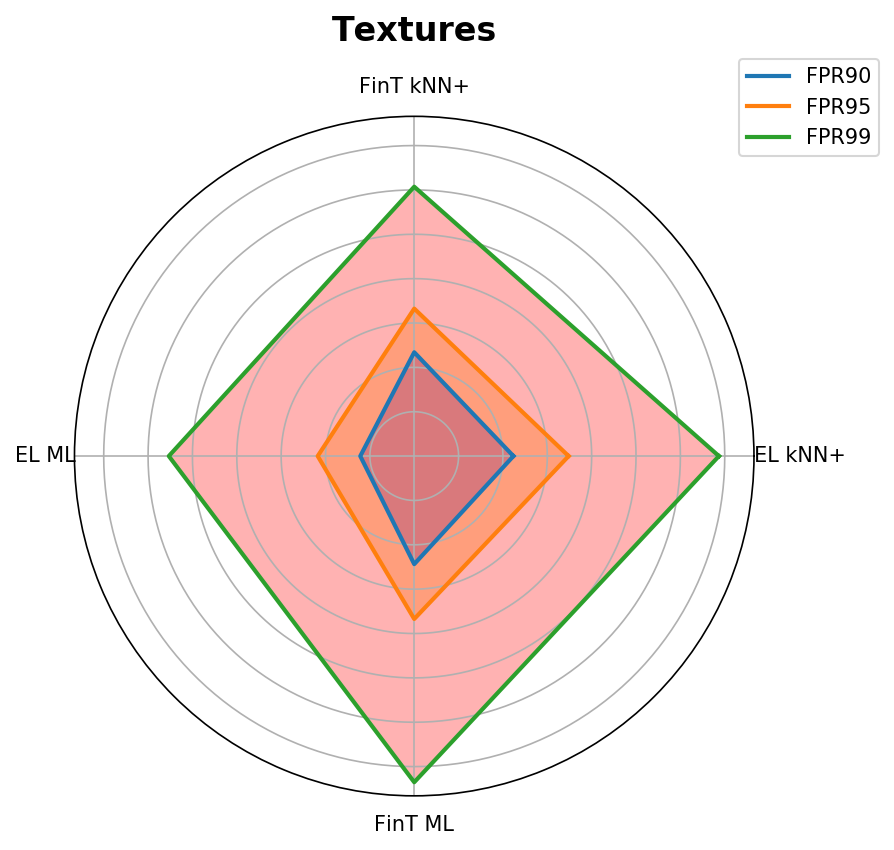}      
    \end{subfigure}
    \caption{Radar plots of false positive rates (FPR) at 90 percent, 95 percent, and 99 percent thresholds for Out-of-Distribution (OOD) detection methods in the informed prior setting.}
\label{fig:4}
\end{figure*}

\subsection{CIFAR-10 Experiments}
Table \ref{Table2} shows that BNNs and empirical Bayes marginally outperform transfer learning on classical NNs in softmax, maximum logit, and our logit-based k-NN scores on average, with some caveats. Noticeably, predictive entropy systematically outperforms softmax entropy, but mutual information exhibits significantly worst performance, an observation that we believe is attributed to the choice of prior distribution (diagonal Gaussian) and we discuss this in more detail in the supplementary material.
We notice the logit methods tend to perform better compared to softmax scores for both BNNs and FinT, but the difference is less substantial. This could suggest that the softmax scores are resource-friendly substitutes in the small-scale setting. In the SVHN and CIFAR-100 tasks, we observed that the vector-based kNN+ methods performed consistently better on the FPR90, FPR95,and FPR99 metrics, as shown in the radar plots of Figure \ref{fig:4}. However the maximum expected logit performed better on Places365 and Textures. 

\begin{table*}[h!]
\centering
\begin{adjustbox}{max width=\textwidth}
\begin{tabular}{lcc|lcc}
\toprule
\multicolumn{3}{c|}{\textbf{MNIST (N=1000)}} & \multicolumn{3}{c}{\textbf{CIFAR-10 (N=5000)}} \\
\midrule
\textbf{Method} & \textbf{AUC-ROC ↑} & \textbf{FPR95 ↓} & \textbf{Method} & \textbf{AUC-ROC ↑} & \textbf{FPR95 ↓} \\
\midrule
SE       & 73.26 (7.66) & 73.69 (18.85) & SE (FinT)  & 90.47 (4.83) & 33.94 (13.38) \\
MI       & 90.39 (4.13) & 33.45 (15.09) & MI         & 68.66 (10.15) & 50.03 (6.72) \\
PE       & 91.78 (5.37) & 29.68 (20.07) & PE         & 91.21 (4.98) & 32.83 (13.39) \\
MLE ML   & 50.43 (14.37) & 87.03 (5.13) & FinT ML    & 90.69 (4.75) & 33.31 (14.50) \\
EL ML    & 93.52 (4.44) & 23.33 (15.91) & \cellcolor{lightgray} EL ML      & $ \cellcolor{lightgray} \mathbf{92.58}$ (4.81) &  \cellcolor{lightgray}$\mathbf{28.03}$ (14.74) \\
MLE kNN  & 85.37 (10.51) & 43.72 (27.67) & FinT kNN   & 81.80 (2.37) & 42.34 (13.26) \\
EL kNN   & 86.51 (4.25) & 41.27 (23.91) & EL kNN     & 81.25 (2.02) & 39.12 (10.99) \\
MLE kNN+ & 87.01 (7.64) & 45.12 (23.15) & FinT kNN+  & 90.28 (4.30) & 31.73 (14.17) \\
\cellcolor{lightgray} EL kNN+ & $\cellcolor{lightgray} \mathbf{93.83}$ (3.67) & $\cellcolor{lightgray}\mathbf{20.20}$ (11.82) & EL kNN+    & 90.88 (4.05) & 29.67 (12.12) \\
\bottomrule
\end{tabular}
\end{adjustbox}
\caption{Average OOD performance, summarising Tables 1 and 2. Left section shows MNIST (N=1000), right section shows CIFAR-10 (N=5000). Each entry reports mean ± standard deviation over the $4$ respective OOD datasets.}
\label{tab:summary_average}
\end{table*}



\section{Discussion}
In this work, we explore the empirical advantages of Bayesian Neural Networks in the context of OOD detection on small training sets. Our detailed results are in Tables \ref{Table1} and \ref{Table2}, and are summarized in Table \ref{tab:summary_average}. Perhaps most notably, the results suggest that Bayesian approaches offer a tangible advantage for OOD detection, particularly when training data is scarce, highlighting  the value of uncertainty-aware models in low-data regimes. The proposed class-conditional kNN-based logit score further illustrates how combining aleatoric and epistemic components in the logit space can yield a more nuanced view of uncertainty, potentially bridging gaps left by existing methods. The superior performance of select Bayesian logit-based OOD scores over predictive entropy and mutual information indicates that the representation space itself may carry richer signals for distinguishing ID and OOD inputs than predictive distributions alone. This points toward a promising direction for future research on BNNs, where leveraging internal representations could provide more robust and scalable tools for uncertainty quantification.

We demonstrated the significance of incorporating prior knowledge, a staple feature of BNNs, when limited training data is available. When no prior model weights are assumed to be known, we found a Gaussian scale-mixture prior provides sufficient flexibility to learn the posterior distribution, and better detect OOD samples. In the setting where prior model weights are known, we implemented and validated that the empirical Bayesian method MOPED achieves top performance compared to the competing method. 
We encourage future works to further investigate the Bayesian advantage by adapting novel OOD scores, and experimenting with Bayesian inference techniques like the Laplace approximation \cite{daxberger2021laplace}.

\bibliography{OOD2024}
\clearpage
\section{Supplementary Material}

\paragraph{Bayesian LeNet-5} We adapt the PyTorch Bayesian CNN implementation by Shridhar et al. \cite{shridhar2019comprehensive}, which efficiently applies the Bayes by Backprop method of Blundell et al. \cite{blundell2015weight} to CNNs 
for the experiments. For MNIST experiments, we follow Shridhar et al. and use the Softplus activation with $\beta=1$ on the LeNet-5 \cite{lecun1998gradient} CNN architecture. 
The variational posterior is taken as a diagonal Gaussian distribution with individual weights $\omega_i \sim \mathcal{N}(\omega_i;\mu_i, \sigma_i^2)$, where $\sigma_i = \log(1 + exp(\rho_i))$, and $\mu_i,\rho_i$ are initialized from $\mathcal{N}(0,0.01)$, $N(-5,0.01)$ respectively. Gradient descent is perfomed on the variational parameters $\theta = (\mu, \rho)$ via the Adam Optimizer with an initial learning rate $0.001$ on batches of size $256$ for a total of $200$ epochs. From mini-batches, we estimate the expected log- likelihood term from the variational objective with the stochastic gradient variational Bayes SGVB estimator\cite{kingma2015variational} and the analytical divergence term is multiplied by $0.1$ for relative weighting. The Gaussian mixture prior hyperparameters were set to $\pi = 0.75$, $\sigma_1=0.1$, and $\sigma_2=0.5$. At each epoch, variational parameters are saved when the validation accuracy improves on the hold-out validation set.

\paragraph{Bayesian ResNet18} For the Bayesian ResNet18, gradient descent is performed on the variational parameters $\theta = (\mu, \rho)$ via the Adam Optimizer with an initial learning rate $0.0001$ on batches of size $32$ for a total of $200$ epochs. Likewise, we estimate the expected log-likelihood term from the variational objective with the stochastic gradient variational Bayes SGVB estimator \cite{kingma2015variational}.  Similar to MNIST training, variational parameters are saved when the validation accuracy improves on a hold-out validation set.

\paragraph{SGD and FinT.} For NN training, we use the Adam optimizer with a learning rate of $0.001$, and train on batches of size $256$ for $200$ epoch. In fine-tuning experiments we use a learning rate of $0.0005$ with Adam optimizer and train on batches of size $32$ with a cross entropy loss. For comparability we train a standard ResNet18 on the CIFAR10 data and a LeNet5 for the MNIST data.

\paragraph{Training split.} We trained the models on the In-Distribution dataset by randomly dividing the training set into an effective training set comprising $80$ percent of data and the remaining $20$ percent constituting the hold-out validation set.

\paragraph{Computational complexity.}In general, Bayesian methods introduce a multiplicative factor of M (the number of MC samples) that increases computational cost, while non-Bayesian methods typically only depend on the number of training samples and the dimensionality of the input data. The range of inference times for non-Bayesian methods is $0.002-0.03$s, and $0.03-0.12$s for Bayesian methods.

\paragraph{Additional Experimental Details.}Following Daxberger et al. \cite{daxberger2020bayesian}, we randomly sample $5000$ In-Distribution test inputs from the In-Distribution test set and $5000$ OOD inputs from each of the OOD datasets. For evaluation the positive class is treated as OOD. It is important to note that there is some disagreement in recent literature on which class to designate as positive; for example \cite{sun2022out, vernekar2019out} define In-Distribution samples to be positive, the opposite convention to the one presented.
Sampling was generally carried out to preserve an equal representation of In-Distribution and OOD labels. Notable instances are the Omniglot dataset, wherein we considered a random selection of $250$ from the $1623$ labels and Places365 wherein we randomly sampled the images. Nguyen et al. \cite{nguyen2022out} remarks that taking fewer samples choice provides a more realistic assumption as it is generally be unfeasible to observe OOD data of the same size, providing a motivation for the designated sampling. We chose k=5 and k=10 to be the k-values for k-NN methods in the MNIST and CIFAR10 experiments respectively, but found consistent experimental performance over a range of k-values.

\paragraph{Impact of Prior Choice on Uncertainty Measures in BNNs} 
Empirical evidence highlights the significant influence of prior distribution choice on the comparative performance of epistemic uncertainty measures, specifically mutual information (MI), in Bayesian neural networks. In CIFAR-10 experiments employing a diagonal Gaussian prior, predictive entropy (PE) consistently outperformed MI for Out-of-Distribution detection, consistent with findings that simpler unimodal priors may limit epistemic uncertainty representation \cite{silvestro2020impact, henning2021bayesian}. Conversely, in MNIST experiments using a more expressive Gaussian mixture prior, MI’s OOD detection performance improved markedly, often matching that of PE. This aligns with Malinin et al.~\cite{malinin2018predictive}, who showed that richer prior structures, such as in Prior Networks, enhance uncertainty calibration and OOD detection by better capturing distributional uncertainty. These results suggest that flexible, multimodal priors enable improved posterior uncertainty modeling, enhancing MI’s reliability as an epistemic uncertainty metric. Simpler priors may fail to capture the complex posterior geometry needed for accurate MI estimation, leading to PE’s superior performance in some settings. Overall, these findings underscore the critical role of prior expressivity in calibrating uncertainty and modulating the relative effectiveness of MI versus PE for OOD detection.


\subsection{KL Divergence Between a Gaussian Mixture and a Single Gaussian}

Let \( p(x_1, \dots, x_K) \) be the joint distribution of independent univariate Gaussian mixtures, where each \( x_k \) is distributed according to a mixture of two univariate Gaussians. The probability density function for \( p(x_k) \) is given by: \(p(x_k) = \pi_1 \, \mathcal{N}(x_k \mid \mu_1, \sigma_1^2) + \pi_2 \, \mathcal{N}(x_k \mid \mu_2, \sigma_2^2)\) where \( \pi_1 + \pi_2 = 1 \), and \( \mathcal{N}(x_k \mid \mu_i, \sigma_i^2) \) denotes a univariate Gaussian distribution with mean \( \mu_i \) and variance \( \sigma_i^2 \) for the \( i \)-th component. We are interested in the Kullback--Leibler (KL) divergence between the multivariate diagonal Gaussian \( q(x_1, \dots, x_K) \) and the joint mixture distribution \( p(x_1, \dots, x_K) \), where each component of \( q \) follows a univariate Gaussian with parameters \( \mu_k^{(q)} \) and \( \sigma_k^{(q)} \): \(q(x_1, \dots, x_K) = \prod_{k=1}^{K} \mathcal{N}(x_k \mid \mu_k^{(q)}, (\sigma_k^{(q)})^2),
\) where \( \mathcal{N}(x_k \mid \mu_k^{(q)}, (\sigma_k^{(q)})^2) \) is a univariate Gaussian with mean \( \mu_k^{(q)} \) and variance \( (\sigma_k^{(q)})^2 \) for each \( x_k \). Since the components \( x_1, \dots, x_K \) are independent, the KL divergence between \( q(x_1, \dots, x_K) \) and \( p(x_1, \dots, x_K) \) can be decomposed as: \(
\mathrm{KL}(q \,\|\, p) = \sum_{k=1}^{K} \mathrm{KL}\!\left(q(x_k) \,\|\, p(x_k)\right).
\) Next, we compute the KL divergence for each \( x_k \), which corresponds to the divergence between the Gaussian \( q(x_k) \) and the mixture distribution \( p(x_k) \). 
Because \( p(x_k) \) is a mixture of two Gaussians, there is no analytic form for this divergence; however, by applying Jensen's inequality, we obtain a tractable upper bound:
\[
\mathrm{KL}\!\left(q(x_k) \,\|\, p(x_k)\right) 
\leq 
\sum_{i=1}^2 \pi_i \, 
\mathrm{KL}\!\left( 
\mathcal{N}(x_k \mid \mu_k^{(q)}, (\sigma_k^{(q)})^2) 
\,\big\|\, 
\mathcal{N}(x_k \mid \mu_i, \sigma_i^2)
\right).
\]

We apply this upper bound as a fast approximation in our MNIST experiments, since the Gaussian components in \( p \) share the same mean (zero) and have slightly differing variances.

The KL divergence between two univariate Gaussians \( \mathcal{N}(\mu_m, \sigma_m^2) \) and \( \mathcal{N}(\mu_n, \sigma_n^2) \) (in the direction \( \mathrm{KL}(\mathcal{N}_m \,\|\, \mathcal{N}_n) \)) is given by:
\[
\mathrm{KL}\!\left( 
\mathcal{N}(\mu_m, \sigma_m^2) 
\,\|\, 
\mathcal{N}(\mu_n, \sigma_n^2)
\right)
= 
\log\!\left( \frac{\sigma_n}{\sigma_m} \right)
+ 
\frac{\sigma_m^2 + (\mu_m - \mu_n)^2}{2 \sigma_n^2}
- 
\frac{1}{2}.
\]

Hence, the upper-bound approximation for the total KL divergence between the joint diagonal Gaussian \( q(x_1, \dots, x_K) \) and the mixture \( p(x_1, \dots, x_K) \) is:
\[
\mathrm{KL}(q \,\|\, p)
\approx
\sum_{k=1}^{K} 
\sum_{i=1}^2 
\pi_i
\left[
\log\!\left( \frac{\sigma_i}{\sigma_k^{(q)}} \right)
+
\frac{(\sigma_k^{(q)})^2 + (\mu_k^{(q)} - \mu_i)^2}{2 \sigma_i^2}
-
\frac{1}{2}
\right].
\]

This expression provides a computationally efficient upper-bound estimate of the KL divergence between a joint diagonal Gaussian and a product of independent Gaussian mixture distributions.

\subsection{Tests of Normality}

We evaluated the univariate normality of the true-class logit values across MNIST training samples using the Shapiro--Wilk test. For each digit class (0--9), we collected 500 samples per image from 1000 images, yielding 500{,}000 logit values per class (selecting the logit dimension that corresponds to the ground-truth label). Due to sample size limitations of the test, we randomly subsampled 5000 values per class. The results revealed statistically significant deviations from normality across all classes ($p$-values $< 5 \times 10^{-5}$), with W-statistics ranging from 0.9830 (class 6) to 0.9984 (class 0). While some classes (e.g., class 0) exhibited only mild deviations, others (e.g., classes 3 and 6) showed more pronounced non-normal behavior. These findings suggest that even when focusing solely on the logit dimension corresponding to the true label, the distribution is not strictly Gaussian.

We performed the Henze-Ziegler (HZ) multivariate normality test to assess whether the 10-dimensional logit vectors, generated by the neural network for each image in the MNIST training set, follow a multivariate normal distribution. Each image was tested with 3000 samples. The HZ test statistic ranged from 1.06 to 2.10, with most values clustered between 1.2 and 1.4, and a few outliers reaching up to 2.1. The corresponding p-values were extremely small ($p \approx 0$ for all cases), indicating strong rejection of the null hypothesis of normality in all cases.The mean HZ statistic across all tests was 1.31, with a standard deviation of 0.14, confirming that the 10-dimensional logit distributions deviate significantly from multivariate normality.

These results suggest that the logit vectors for each image do not follow a multivariate normal distribution, reflecting the complex, non-linear transformations inherent in neural network architectures. This finding is consistent with common observations in deep learning, where model outputs, such as logits, tend to exhibit deviations from normality due to the influence of activation functions and the network’s design.

The Henze-Zirkler multivariate normality tests were performed for each class (0 through 9) by grouping images based on their labels, using 500 samples per image. The results show a strong rejection of the normality hypothesis for all classes, as indicated by the very small p-values (approaching 0 for each class) and the relatively high HZ statistics for each test.

The results from the normality tests for the true label logits across 1000 images show that only 2 images (0.20 percent) exhibit normality in their respective logits corresponding to the true class value. 

\subsection{Logit distribution complexity}

We investigated the structure of the stochastic logit distributions produced by our model under repeated sampling (500 samples per image). First, we conducted Hartigan's dip test on each dimension of the logits individually. The results indicated that 100\% of the univariate logit dimensions across all images were statistically consistent with unimodality (p-value $> 0.05$).

However, when evaluating the multivariate distribution of the 10-dimensional logit vectors per image, we observed substantial complexity. Using KMeans clustering with $k=2$ and measuring the relative within-cluster variances, the majority of images were flagged as potentially multimodal. This suggests that, while individual logit dimensions appear unimodal, the joint distribution across dimensions is often multimodal or structured in complex ways. Interestingly, despite this internal complexity, we found that the mean logit vector per image provided a strong and robust feature for OOD detection. This observation is consistent with the idea that the first moment (mean) of the logit distribution captures the dominant shift between in-distribution (ID) and Out-of-Distribution (OOD) samples, even when higher-order structure (e.g., multimodality) is present within the sample set.

This explains why full-distribution-based methods, such as Wasserstein distance applied across all samples, struggled to clearly distinguish ID from OOD, whereas simple metrics based on the mean logits performed substantially better. The mean logit acts as a stable and informative representation, effectively suppressing internal stochastic variation and highlighting overall distributional shifts.

\subsection{Additional Plots}
\begin{figure*}[!h]
 \includegraphics[scale=0.28]{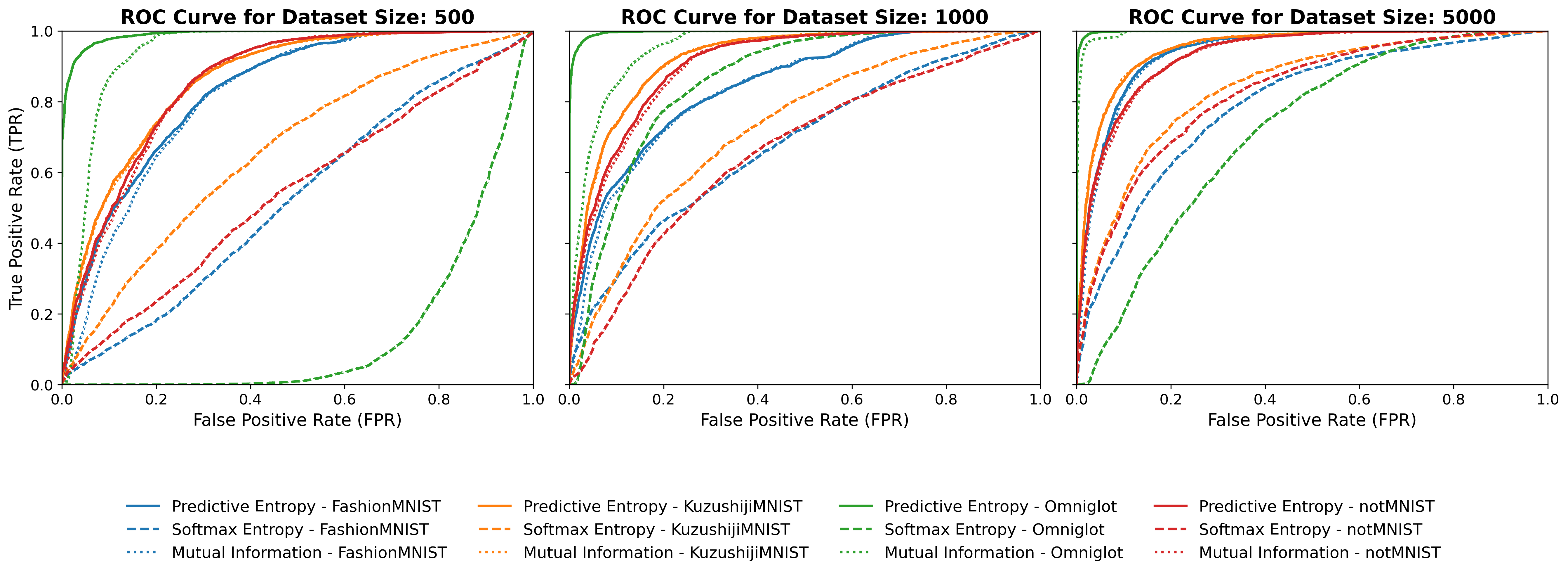}
 \caption{ROC curves comparing OOD detection performance across three dataset sizes (1000, 5000, 10000) for four datasets (FashionMNIST, KuzushijiMNIST, Omniglot, notMNIST). Three uncertainty estimation methods—Predictive Entropy, Softmax Entropy, and Mutual Information—are evaluated. Each plot shows how performance varies with dataset size, with distinct line styles representing different methods. A higher curve signifies better discrimination between In-Distribution and OOD data}
 \label{fig:7}
 \end{figure*}

\begin{figure*}[h]
    \centering
    \includegraphics[scale=0.30]{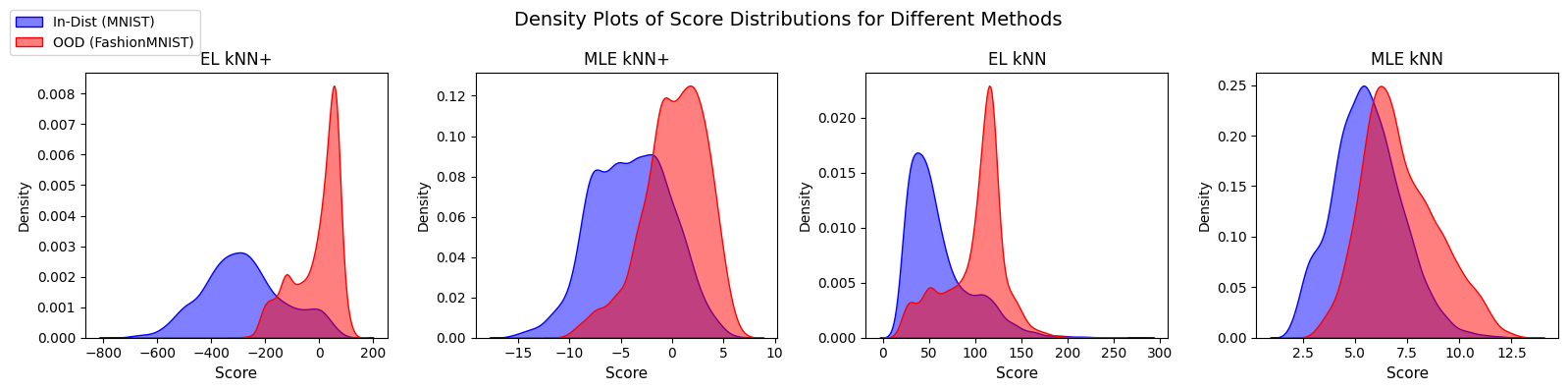}
    \caption{Density plots showing the score distributions of In-Distribution (MNIST, blue) and Out-of-Distribution (FashionMNIST, red) samples across different kNN-based methods: EL kNN, MLE kNN, EL kNN+, and MLE kNN+. The plots illustrate how the score distributions for each method compare between the two datasets.}
    \label{fig:enter-label}
    \label{fig:8}
\end{figure*}

\begin{figure*}[!t]
    \centering
    \includegraphics[scale=0.40]{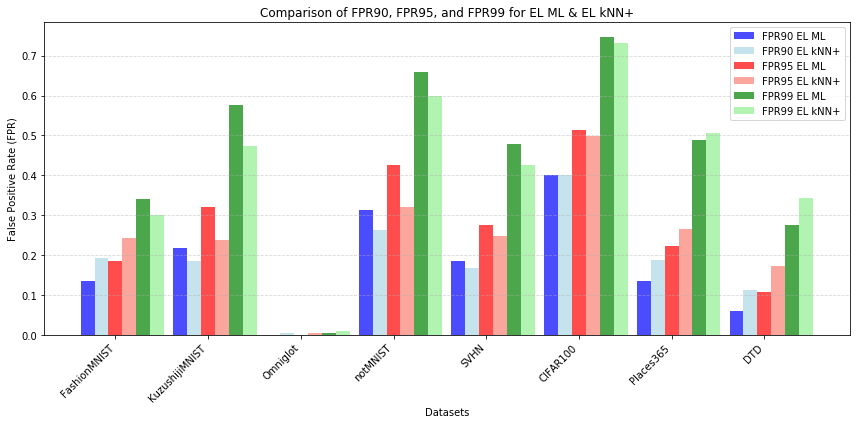}
    \caption{Comparison of false positive rates (FPR) at 90 percent, 95 percent, and 99 percent thresholds for Out-of-Distribution (OOD) detection methods. The FPR values are shown for the EL ML and EL kNN+ methods across multiple datasets, including FashionMNIST, Omniglot, KuzushijiMNIST, and others. The bars represent the FPR at different thresholds, providing insight into the performance of each method at varying levels of stringency.}
    \label{fig:9}
\end{figure*}

This section provides additional plots to supplement the experimental findings. Figure \ref{fig:7} extends Figure \ref{fig:1}  to all MNIST experiment sizes. Figure \ref{fig:8} shows density plots of k-NN based scores, visually depicting improved separability by our class-conditional score. Figure \ref{fig:9} provides additional FPR data for the informed prior CIFAR10 experiments.


\end{document}